\documentclass[letterpaper, 10 pt, conference]{ieeeconf}  

\IEEEoverridecommandlockouts                              

\usepackage{amsmath} 
\usepackage{amssymb}  
\usepackage{graphicx}
\usepackage{mathtools}
\usepackage{subfig}
\usepackage{adjustbox}
\usepackage[noadjust]{cite}
\usepackage{lipsum}
\usepackage{multirow}
\usepackage{graphicx}
\usepackage{amssymb,mathtools}
\usepackage{subfig}
\usepackage{multirow}
\usepackage{hhline}
\usepackage{diagbox}
\usepackage{hhline}
\usepackage{etoolbox}
\usepackage{capt-of}
\usepackage{caption}
\usepackage{multicol}

\title{\LARGE \bf
Constructing Category-Specific Models for Monocular Object-SLAM
}

\author{Parv Parkhiya$^{1}$, Rishabh Khawad$^{1}$, J. Krishna Murthy$^{1}$, Brojeshwar Bhowmick$^{2}$, and K. Madhava Krishna$^{1}$
\thanks{$^{1}$ Parv Parkhiya, Rishabh Khawad, J. Krishna Murthy, and K. Madhava Krishna are with the Robotics Research Center, KCIS, IIIT Hyderabad, India. {\tt\small albert.author@papercept.net}}
\thanks{$^{2}$ Brojeshwar Bhowmick is with Tata Consultancy Services, Kolkata, India.}
}

\begin{document}

\maketitle
\thispagestyle{empty}
\pagestyle{empty}

\begin{abstract}

We present a new paradigm for real-time object-oriented SLAM with a monocular camera. Contrary to previous approaches, that rely on \emph{object-level} models, we construct \emph{category-level} models from CAD collections which are now widely available. To alleviate the need for huge amounts of labeled data, we develop a rendering pipeline that enables synthesis of large datasets from a limited amount of manually labeled data. Using data thus synthesized, we learn category-level models for object deformations in 3D, as well as discriminative object features in 2D. These category models are instance-independent and aid in the design of object landmark observations that can be incorporated into a generic monocular SLAM framework. Where typical object-SLAM approaches usually solve only for object and camera poses, we also estimate object shape on-the-fly, allowing for a wide range of objects from the category to be present in the scene. Moreover, since our 2D object features are learned discriminatively, the proposed object-SLAM system succeeds in several scenarios where sparse feature-based monocular SLAM fails due to insufficient features or parallax. Also, the proposed category-models help in object instance retrieval, useful for Augmented Reality (AR) applications. We evaluate the proposed framework on multiple challenging real-world scenes and show --- to the best of our knowledge --- first results of an instance-independent monocular object-SLAM system and the benefits it enjoys over feature-based SLAM methods.

\end{abstract}

\section{INTRODUCTION}

Simultaneous Localization and Mapping (SLAM) finds various real-world applications such as autonomous navigation, visual inspection, mapping, and surveillance. Monocular cameras have evolved as popular choices for SLAM, especially on platforms such as hand-held devices and Micro Aerial Vehicles (MAVs). Most state-of-the-art monocular SLAM systems \cite{ORB} operate on geometric primitives such as points, lines, and planar patches. Others operate directly on images, without the need for expensive feature extraction steps \cite{LSD}. However, both these sets of approaches lack the ability to provide a rich semantic description of the scene.

Recognizing and keeping track of objects in a scene will enable a robot to build meaningful maps and scene descriptions. Object-SLAM is a relatively new paradigm \cite{Sunderhauf,SLAM++,Paull} towards achieving this goal. Summarized in one sentence, object-SLAM attempts to augment SLAM with object information so that robot localization, object location estimation (in some cases, object pose estimation too), and mapping are achieved in a unified framework.

There are two dominant paradigms in object-SLAM research, depending on the way objects are characterized in the SLAM framework. In the first paradigm \cite{SLAM++,Choudhary}, \emph{object-level} (instance-specific) models are assumed to be available beforehand. However, the very nature of monocular SLAM with scale ambiguity coupled with the loss in information due to projection onto the image plane renders this paradigm infeasible for monocular object-SLAM systems. The second paradigm \cite{Sunderhauf2015,SfM++}, assumes a \emph{generic model}, regardless of the object category. For instance, \cite{SfM++} models all objects as ellipsoids, and \cite{RAS2016,Paull} model all objects as cuboids. Both these approaches suffer a few disadvantages. Relying on object-level models will result in the need to have precise object models for all instances of an object category. On the other hand, generic models do not give much information about an object beyond the object category label. In many applications, such as manipulation for instance, it is advantageous to know the object pose.

\begin{figure}[!t]
	\centering 
	\includegraphics[scale=0.1]{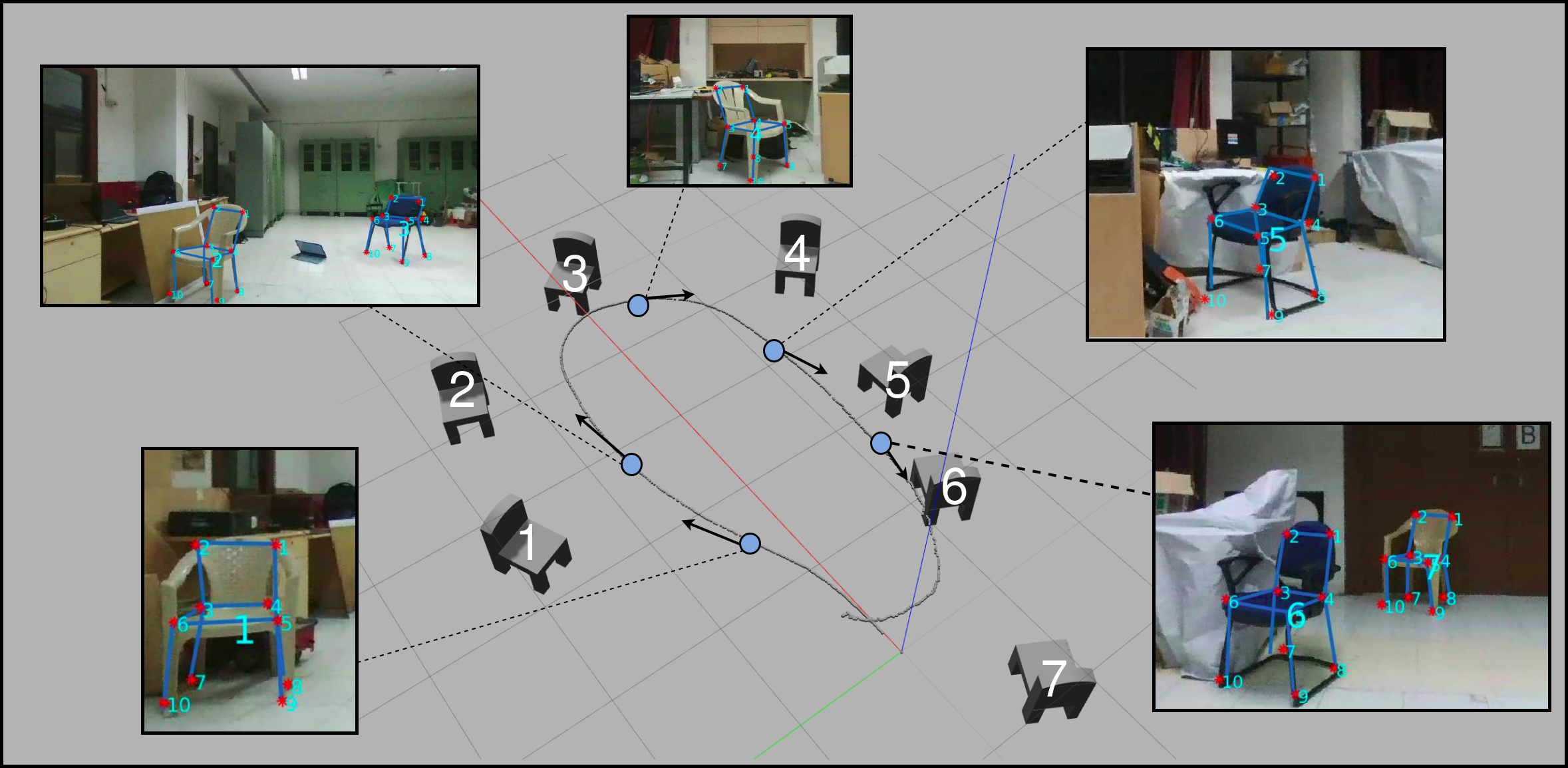}
	\caption{Sample output of the proposed object-SLAM system. Given RGB images from a monocular camera (here from a quadrotor), we estimate the trajectory of the camera, as well as the poses and shapes of various objects in the scene. The proposed \emph{category-specific models} can be used either in an incremental setting for online SLAM, or in a batch setting for offline factor-graph optimization.}
	\label{fig:teaser}  
	\vspace{-0.75cm}
\end{figure}

In this paper, we propose a new paradigm for monocular object-SLAM, that combines the best-of-both-worlds. To enjoy the expressive power of instance-specific models yet retain the simplicity of generic models, we construct \emph{category-specific} models, i.e., the object category is modeled as a whole. We employ the widely used linear subspace model \cite{KM_ICRA,Zia,tulsiani_PAMI} to characterize an object category and define object observations as factors in a SLAM factor graph \cite{ISAM2,RiSE}. In our object-SLAM formulation, we do not assume any knowledge about the instance (interchangeably referred to as \emph{shape}) of the object. Rather, we explicitly solve for the object shape in a joint formulation. The object-SLAM backend estimates robot trajectory and map, as well as poses and shapes of all objects in the scene.

Naturally, one would expect that a lot of data will be needed to learn category-specific models that generalize well across object instances, and rightly so. Datasets such as ShapeNet, SceneNet, ObjectNet have made available CAD collections of various object categories. We exploit the ready availability of such CAD collections to construct our category models. These category models capture the deformation modes of objects in 3D. Correspondingly, we leverage recent successes of Convolutional Neural Networks (CNNs) in keypoint localization \cite{KM_ICRA,KM_IROS,kostas_ICRA,hourglass} to train 2D object feature extractors. To alleviate the need for large amounts of manually annotated training data, we design a \emph{render} pipeline, along the lines of RenderForCNN \cite{RenderForCNN}, to synthesize enormous amounts of training data for category model learning. The presented render pipeline takes in a small volume of manually annotated data and synthesizes a large dataset that can be used to efficiently train a 2D object feature extraction network. We show that feature detectors learned from the render pipeline are more precise compared to ones that are learned over real data alone, which corroborates claims in \cite{RenderForCNN}.

We evaluate our object-SLAM system on multiple challenging real-world sequences and present --- to the best of our knowledge --- first steps towards instance-independence in monocular object-SLAM. Since we use discriminative 2D features on objects, our system is robust to conditions such as strong rotation, where monocular SLAM approaches usually face catastrophic failure. We present both incremental and batch versions of our object-SLAM pipeline and demonstrate its advantages qualitatively and quantitatively over feature-based visual SLAM approaches \cite{ORB}. Finally, we show that, using our category-level models, one can perform object instance retrieval and this can be used in many Augmented Reality (AR) applications for overlaying object models in a scene. Fig. \ref{fig:teaser} illustrates an output from our pipeline. Objects are consistently embedded onto the robot's trajectory and their 3D models are rendered.

\begin{figure*}[!htb]
\centering 
\includegraphics[width = \textwidth]{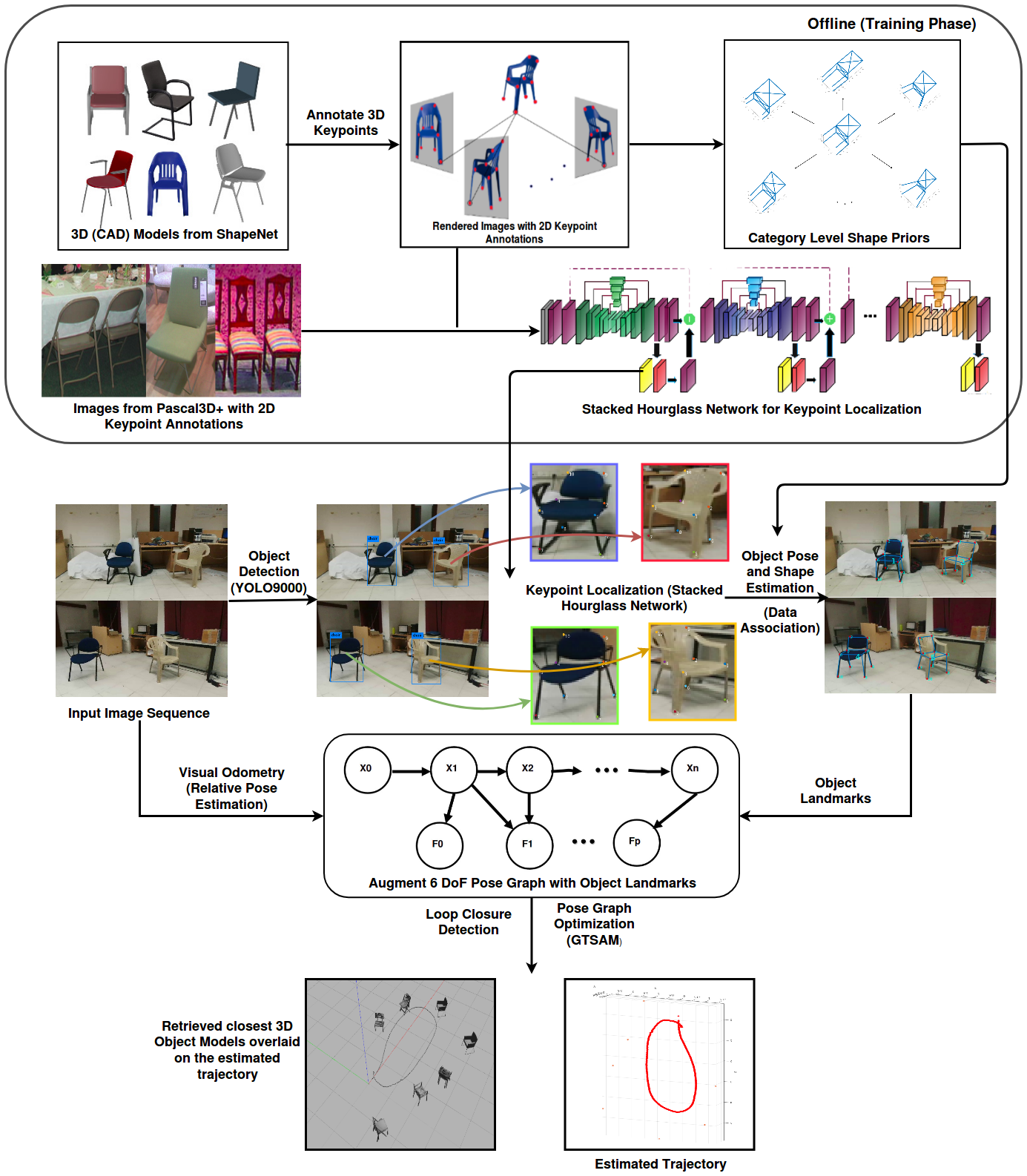}
\caption{Complete pipeline for learning and incorporating category-level models for object-SLAM}
\label{fig:pipeline}  
\end{figure*}

\section{Related Work}

Nearly all state-of-the-art SLAM systems \cite{ORB,LSD,SVO} rely on pose-graph (or other factor graph) optimization \cite{g2o,GTSAM}. In this section we review related work on object-SLAM, and outline certain limitations in them that form motivating factors for the proposed approach.

\subsection{Object-SLAM}

With recent advances and subsequent stabilization of SLAM systems, the community has been devoting attention to incorporating objects into the SLAM framework. To this end, some recent approaches for object-oriented SLAM have been proposed \cite{SLAM++,Paull,RAS2016,SfM++,Sunderhauf,Sunderhauf2015}.

Most of these works rely on depth information from RGB-D or stereo sensors \cite{SLAM++,Choudhary,Paull,Sunderhauf2015}. In \cite{SLAM++,Choudhary}, assume an instance-level model of the objects are known a priori. In \cite{SLAM++}, a real-time 3D object detection algorithm is applied on an RGB-D image stream and these objects are fused along with odometry information in a pose graph optimization scheme. Similarly in \cite{Choudhary}, a framework for multi-robot object-SLAM is proposed. Again, each robot is equipped with an RGB-D sensor and object models are available a priori.

There is another paradigm in which no instance-level models are available a priori. In \cite{Paull}, association and object poses are solved for jointly, in a factor graph framework, using data from RGB-D cameras. Among monocular object-SLAM/SfM approaches, \cite{SfM++,RAS2016} fall under this paradigm. In such approaches, objects are modeled as bounding boxes \cite{RAS2016,Sunderhauf2015} or as ellipsoids \cite{SfM++}.

Our method hence falls under a \emph{third} paradigm, where we assume category-models, and not instance-level models.

\subsection{Object-Category Models}

Over the last few years, object-category models have been applied to several problems in monocular vision. In \cite{tulsiani_PAMI,KM_ICRA,Zia}, category-level models are employed to obtain object reconstructions from single images. These approaches demonstrate that the loss of information in the monocular imaging process can be compensated for by incorporating priors on shapes of objects belonging to a particular category.

We use these category models and exploit them to design object observation factors that can be easily incorporated into monocular SLAM, and also generalize across several instances from the category, without the need for modeling all possible instances from the category.

\subsection{Keypoint Localization Using CNNs}

Convolutional Neural Networks (CNNs) have been the driving reason behind recent advances in object detection \cite{YOLO,Fast-RCNN} and object keypoint localization \cite{hourglass,VpsKps,kostas_ICRA,KM_IROS}. When run on a GPU, these CNNs are capable of processing image frames with latencies of about 100-300 milliseconds and form important components of our pipeline.

\subsection{Render Pipelines for Data Synthesis}

With the advent of CAD model collections such as \cite{Shapenet}, 3D data is now in abundance. In \cite{RenderForCNN}, image synthesis using a rendering engine is proposed as an alternative to training on real images annotated manually. Models trained for the task of object viewpoint prediction on rendered data (and subsequently finetuned on a smaller dataset comprising real data) are shown to outperform models that have been trained on (larger) real datasets alone. Our experiments corroborate this fact for the task of object keypoint prediction as well.

We build on several of the components described here, however we craft the outputs to create object factors that can be augmented to factor graphs \cite{ISAM2} constructed using monocular SLAM approaches. The entire pipeline is summarized in Fig. \ref{fig:pipeline} and is explained in the subsequent sections.


\section{Constructing Category-Specific Models}

In this section, we describe the category-level models that we employ in our object-SLAM system. Our key insight is that shapes of objects from the same category are not \emph{arbitrary}. Instead they all follow a set pattern that can be learned by examining several instances. We adopt the widely used linear subspace model \cite{Zia,tulsiani_PAMI,KM_ICRA,KM_IROS} and represent objects as 3D wireframes.

\subsection{Category-Level Model}

In the proposed approach, we lay an emphasis on the use of \emph{category-level} models as opposed to \emph{instance-level} models for objects. To construct a category level model, each object is first characterized as a set of 3D locations of salient points that are common across all instances of the category. For example, such salient points for the \emph{chair} category could be legs of the chair, corners of the chair backrest, and so on. These points are common across all instances of the category. We denote by $\mathbf{S}$ the $3K$-vector comprising of an ordered collection of 3D locations of these $K$ salient points, which we refer to as \emph{keypoints} for that object category. Keypoints for the \emph{chair} category which we use throughout this work, are shown in Fig. \ref{fig:pipeline}.

According to the linear subspace model \cite{KM_ICRA}, the space of actual shapes of a category are confined to a much lower dimensional subspace of $\mathbb{R}^{3K}$. This is easy to see, since there are several dependencies that exist among the $K$ keypoints, such as subsets of the keypoints being planar, symmetric, etc. Any object can then be expressed as a \emph{mean shape} that can be deformed along linearly independent directions (\emph{basis vectors}) using \emph{shape parameters} (coefficients of the linear combination).

Mathematically, if $\mathbf{\bar{S}}$ is the mean shape of the category, and $\mathbf{V_i}$s are a deformation basis obtained from PCA over a collection of aligned ordered 3D CAD models (such as \cite{Shapenet}),
\begin{equation}
\mathbf{S} = \mathbf{\bar{S}} + \sum_{b=1}^{B} \lambda_b \mathbf{V_b} = \mathbf{\bar{S}} + \mathbf{V \Lambda}
\label{eqn:shapePrior}
\end{equation}
where $B$ is the number of basis vectors (the top-$B$ eigenvectors after PCA). Here, $\mathbf{V}$ represents the learnt $3K \times B$ deformation modes.$\mathbf{\Lambda}$ is a $K$-vector containing the deformation coefficients. Varying the deformation coefficients produces various shapes from the learnt shape subspace, as shown in Fig. \ref{fig:pipeline} (panel: \emph{Category Level Shape Priors}).

\subsection{Discriminative Feature Extraction}

Since our category-level models rely on keypoints for that category, we need reliable discriminative feature extractors that localize the corresponding keypoints in 2D images. For the same, we adopt the \emph{stacked hourglass} model introduced in \cite{hourglass} and subsequently modified in \cite{KM_IROS}. \vspace{-0.3cm}

\subsection{Render Pipeline}
\vspace{-0.2cm}
Inspired by RenderForCNN \cite{RenderForCNN}, we implement our customized render pipeline for generating huge amounts of synthetic keypoint annotated chair images using a small set of 3D annotated keypoints. We briefly summarize the steps in our render pipeline, and how we exploit its advantages for learning 3D category-models as well as discriminative 2D feature extractors.

\subsubsection*{\textbf{Render Pipeline for 3D Category-level Models}}
First, we choose a collection of CAD models of chairs, comprising of about 250 chairs sampled from the ShapeNet \cite{Shapenet} repository. For each chair, we synthesize a few (typically 8) 2D images with predetermined viewpoints (azimuth, elevation, and camera-tilt angles). Keypoints in these images are then annotated (in 2D) manually, and then triangulated to 3D to obtain 3D keypoint locations on the CAD model.

Since the models are already assumed to be aligned, performing a Principal Component Analysis (PCA) over the (mean-subtracted) 3D keypoint locations results in the deformation basis (eigenvector obtained from PCA). This constitutes the category-level model learning phase.

\subsection*{\textbf{Render Pipeline for 2D Feature Extraction}}
For training 2D feature extractors, we use the 3D keypoint annotated CAD models to synthesize 2D images with Blender \cite{blender} as the rendering engine. Apart from using the RenderForCNN \cite{RenderForCNN} framework to randomly sample view parameters and generate 2D images with overlaid backgrounds, we also perform a projection of the 3D keypoints to 2D using the same view parameters to obtain 2D keypoint annotations for training the stacked hourglass architecture \cite{hourglass,KM_IROS,kostas_ICRA} for keypoint localization.

\subsection*{\textbf{Advantages of the Proposed Render Pipeline}}
\begin{itemize}
\item Only 2D annotation is required. Annotation on 3D CAD models requires expensive labeling effort. We, however, perform annotations on a small set of rendered 2D images and triangulate them to 3D.
\item Keypoint annotations will be available even for occluded parts. This is one major drawback of keypoint localization architectures trained over real data. Since there are no reliable estimates of keypoint locations for occluded parts, these labels are not present in the ground-truth in datasets such as PASCAL3D+ \cite{PASCAL3D}. We demonstrate that our models perform much better due to the availability of occluded keypoint locations as well.
\item Using only small volumes of labeled data, we can generate millions of training examples for the keypoint localization CNN.
\end{itemize}

An overview of the render pipeline developed is illustrated in Fig. \ref{fig:pipeline}.

\section{Object Measurement Factors Using Category-Models}

In this section, we describe how the category-models thus learned can be incorporated into a monocular SLAM backend. We first introduce the notation we use throughout the section. $\mathbf{T_{ij}} \in SE(3)$ denotes a rigid-body transform that takes a 3D point expressed in the camera frame at time $i$ and expresses it with respect to the camera frame at time $j$. Hence, $\mathbf{T_{ij}} = \mathbf{T_j}\mathbf{T_i}^{-1}$ ($\mathbf{T_*}$ denotes a transformation matrix that transforms points from frame $*$ to the world frame $W$). Each such matrix $\mathbf{T_{ij}}$ is a $4 \times 4$ matrix and takes the following form.
\vspace{-0.65cm}

\begin{equation}
\mathbf{T_{ij}} = \left[ \begin{matrix} \mathbf{R}_{ij} & \mathbf{t}_{ij} \\ \mathbf{0} & 1 \end{matrix} \right] \space  where \mathbf{R}_{ij} \in SO(3), \mathbf{t}_{ij} \in \mathbb{R}^3
\label{eqn:transformMat}
\end{equation}

Each camera frame $i$ is related to camera frame $j$ in the following manner. If $\mathbf{^iX}$ denotes the 3D coordinates of a world point $\mathbf{^wX}$ in the $i$th camera frame, $\mathbf{^jX}$ is given by $\mathbf{^jX} = \mathbf{T_{ij}{^i}X}$. We use $\pi: \mathbb{R}^4 \mapsto \mathbb{R}^2$ that projects 3D homogeneous coordinates to 2D Euclidean coordinates. We subsume the camera intrinsics $f_x, f_y, c_x, c_y$ into the function $\pi$.
\vspace{-0.3cm}

\begin{equation}
\pi \left(\begin{matrix}
X \\ Y \\ Z \\ 1
\end{matrix} \right) = \left( \begin{matrix}
{f_x X \over Z} + c_x \vspace{0.25cm} \\ {f_y Y \over Z} + c_y
\end{matrix} \right)
\end{equation}

We use $Log(.)$ to denote the logarithm map that maps an element of the group $SE(3)$ to a corresponding element in the exponential coordinates of its tangent space $se(3)$.

\subsection{3D Pose-SLAM}

We formally define the 3D pose-SLAM problem as follows. Given a set $\mathcal{Z}$ of \emph{relative} pose measurements $\mathbf{\hat{T}_{ij}}$ among robot poses $i,j$, where $i,j \in \{1..N\}$, estimate $\mathbf{T_i}$ for all $i \in \{1..N\}$ such that the log-likelihood of the observations (relative pose measurements) is maximized. This reduces to the problem of minimizing the observation errors (minimizing the negative log likelihood). We assume that each relative-pose measurement ${\mathbf{\hat{T}_{ij}}}$ has an associated uncertainty $\mathbf{\Sigma_{ij}}$.

\begin{equation}
\underset{\mathbf{T}_i, i \in \{1..N\}}{\text{min}} \space \mathcal{E}_{pose} = \sum_{ {\mathbf{\hat{T}_{ij}}} \in \mathcal{Z} } \| Log({\mathbf{\hat{T}_{ij}}}^{-1} \mathbf{T}_j \mathbf{T}_i^{-1}) \|_{\mathbf{\Sigma_{ij}}}
\label{eqn:SLAM}
\end{equation}

One popular approach of minimizing the error in (\ref{eqn:SLAM}) is the use of factor graphs \cite{ISAM2,GTSAM}. We adopt this approach since it naturally extends to an incremental optimization framework that enables online processing.

\subsection{Object Observation Factors}

Using the linear subspace model illustrated in (\ref{eqn:shapePrior}), we now design object observation factors that are suitable for instance-independent monocular object-SLAM. Given 2D locations of object keypoints $\mathbf{s}$, we use the pose and shape adjustment scheme proposed in \cite{KM_ICRA} to \emph{lift} the 2D keypoints to 3D, thereby estimating the shape and pose of the object from just a single image.

We assume that, from the $i^{th}$ pose, the robot observes $M$ objects ($M$ could also be zero). The set of object observations in frame $i$ is denoted $\mathcal{O}_i$, and each observation in the set is indexed as $\mathbf{\hat{T}^{O_m}_{i}}$. We denote the number of keypoints of the object category by $K$. The $k^{th}$ keypoint of the $m^{th}$ object observed are denoted by $\mathbf{s_k^m}$. The pose and shape of each object can then be computed by minimizing the following keypoint reprojection error (see \cite{KM_ICRA} for details).

\begin{equation}
\underset{ \mathbf{\hat{T}^{O_m}_i}, \mathbf{\Lambda_m} }{\text{min}}
\left(  \sum_{ m = 1 }^M \sum_{k=1}^{K}  \|  \pi(\mathbf{K}\mathbf{\hat{T}}^{O_m}_{i}(\bar{\mathbf{S}} + \mathbf{V \Lambda_m})) - \mathbf{s_k^m}  \|_2^2 \right) + \rho(\mathbf{\Lambda_m})
\label{eqn:poseAdjustment}
\end{equation}

In the above equation, $\rho(\mathbf{\Lambda_m})$ denotes an appropriate regularizers (such as an $L2$ norm, for instance) to prevent shape parameter ($\mathbf{\Lambda}$) estimates from deviating too much from the category-model.

We estimate each object's shape ($\mathbf{\Lambda}$) and pose ($\mathbf{\hat{T}^{O_m}_{i}}$) in each frame $i$ by alternating minimization of the error term in (\ref{eqn:poseAdjustment}), with respect to the pose and shape parameters. If the same object has been associated successfully across multiple frames, we also exploit temporal consistency \cite{KM_IROS} for more precise estimates.

\subsection{Object-SLAM}

The object pose observations arising from (\ref{eqn:poseAdjustment}) now form additional factors in the SLAM factor graph. If $\mathbf{\hat{T}^{O_m}_{i}}$ denotes the pose of object $m$ with respect to the $i^{th}$ camera frame, for each object node in the factor graph, the following pose error is to be minimized.

\begin{equation}
\underset{\mathbf{\hat{T}^{O_m}_{i}}, m \in \{1..M\}}{\underset{\mathbf{T}_i, i \in \{1..N\}}{\text{min}}} \space \mathcal{E}_{obj} = \sum_{i=1}^N \sum_{m=1}^{M} \| Log( \left( \mathbf{\hat{T}^{i}_{O_m}} \right)^{-1} \mathbf{T}_{i}^{-1} \mathbf{T^{O_{\phi(m)}}}  ) \| 
\label{eqn:ObjectSLAM}
\end{equation}

Here, $\mathbf{T^{O\phi(m)}}$ denotes the pose of object $O_{\phi(m)}$ with respect to the global coordinate frame, where	 $\mathbf{\phi(m)}$ denotes the data association function which assigns a globally unique identifier to each distinct object observed so far. The data association pipeline is briefly outlined in Section \ref{sec:implementation}.

Finally, the object-SLAM error that jointly estimates robot poses as well as object poses from relative pose observations is given by \vspace{-0.4cm}

\begin{equation}
\mathcal{E} = \mathcal{E}_{pose} + \mathcal{E}_{obj}
\label{eqn:ObjectSLAM_abstract}
\end{equation}

\section{Implementation Details}
\label{sec:implementation}

We use the publicly available GTSAM \cite{GTSAM} framework for constructing and optimizing the proposed factor graph model. Robot pose observations are obtained from a visual odometry/SLAM approach. We use odometry information published by ORB-SLAM \cite{ORB} for this purpose. However, monocular SLAM inherently suffers from scale ambiguity, i.e., the absolute scale of reconstruction cannot \emph{usually} be recovered. To recover actual scale, we use information from a range sensor to estimate the height of the drone (camera) above the ground. Thereby, using objects that lie on the ground plane (chairs, for instance), we can compute the absolute scale factor by backprojection via the ground plane \cite{KM_ICRA,chandraker2015}. Moreover, ShapeNet \cite{Shapenet} CAD models for the categories we use are available in metric scale. Hence, a rough prior on object dimensions is available and can be used to initialize the solution for faster convergence \cite{KM_ICRA}.
\vspace{-0.2cm}

\subsection{Object Detection and Data Association}
\vspace{-0.2cm}

We use the YOLO9000 \cite{YOLO} object detector to detect objects of interest and perform non-maximum suppression. Each bounding box detection is then passed to the keypoint localization network. 

Between successive frames, objects are tracked using a greedy tracker based on the Hungarian algorithm. For long-term tracking and for Object Loop Closure detection (OLC), we use the estimated shape and pose parameters as costs in the Hungarian algorithm. This has shown to be an extremely simple, yet effective data association method in \cite{Junaid_ICRA}.

The object shape and pose are optimized using Ceres solver \cite{Ceres} and are cast as observations into the factor graph. The factor graph can be optimized in incremental or batch mode, and with or without object loop closures.


\begin{table}[!t]
\centering
\caption{Keypoint localization accuracy (2D) for the stacked hourglass network trained using the proposed render pipeline}
\begin{adjustbox}{width=\linewidth}
\begin{tabular}{|c|c|c|c|}
\hline
\backslashbox{\textbf{Test Accuracy}}{\textbf{Train Accuracy}} & Synthetic only & Real only & Synthetic + Real  \\
 \hline
Synthetic data & $90.67 \%$ & $85.90 \%$ & $\mathbf{94.51 \%}$ \\
\hline  
Pascal3D+ & $75.99 \%$ & $87.46 \%$ & $\mathbf{93.4 \%}$ \\
\hline
Sequence 1 (Ours) & $21.11 \%$ & $66.18 \%$ & $\mathbf{95.93 \%}$ \\
\hline
\end{tabular}

\label{table:keypointAccuracy}
\end{adjustbox}
\vspace{-0.5cm}
\end{table}

\section{Results}

\begin{figure*}[!h]
	\centering 
	{\includegraphics[width = \textwidth]{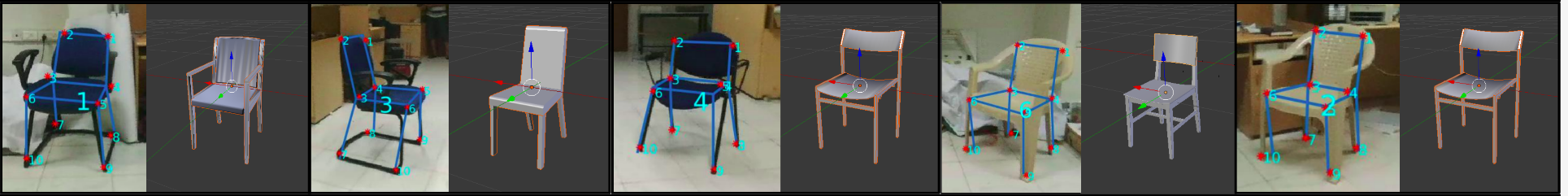}
		\caption{Qualitative results for instance retrieval. Query images and their retrieved CAD models are shown.}
		\label{fig:instanceRetrieval}}  
	\vspace{-0.45cm}
\end{figure*}

\subsection{Object-SLAM Evaluation}
\vspace{-0.1cm}

In this section, we present experimental results upon evaluating various modes of operation of the proposed object-SLAM approach and the render pipeline on multiple real-world sequences. By explicitly modeling objects in a SLAM framework, we achieve substantially lower object localization errors over feature-based SLAM \cite{ORB}. Also, we highlight the use of discriminatively learned features that enable us to obtain trajectory and object pose estimates even in scenarios where monocular SLAM approaches fail, such as strong rotational motion.
\vspace{-0.2cm}

\subsection{Dataset}

To demonstrate object SLAM in a real-world setting, we collected a dataset comprising of 4 monocular video sequences from robots operating in office and laboratory environments. These sequences were collected from a micro aerial vehicle (MAV) flying at a constant height above the floor. For one of the sequences, fiducial markers were placed in the environment and were used to estimate ground truth camera pose and object locations. For two other sequences, no markers were placed in the environment, to deny any undue advantage enjoyed by monocular SLAM. In these sequences, all object locations were measured a priori and this is used in evaluating the object localization accuracy. There was a \emph{rotation-only} run, where the MAV rotates more-or-less \emph{in-place}. In all these runs, we evaluate object localization precision for ORB-SLAM \cite{ORB} and for various flavors of the proposed object-SLAM approach.

A brief description of the dataset is as follows.
\begin{itemize}
\item Sequence 1 : An elongated loop with two parallel sides exhibiting dominant straight motion (7.6 meters) (fiducial markers were used for acquiring ground-truth).
\item Sequence 2 : Smaller run with smoother turns (no fiducial markers).
\item Sequence 3 : 360$^{\circ}$ Rotation in place without any translation from origin.
\item Sequence 4 : 2 meters forward, sharp 90$^{\circ}$ rotation towards left, 4.2 meters forward
\end{itemize}
In all these sequences we consider \emph{chairs} as the objects of interest. For a fair evaluation, and to demonstrate the efficacy of the proposed object-SLAM system, we ensure that most of the frames in the sequence contain various chair instances.

\begin{figure}[!htb]
	\centering
	\includegraphics[scale=0.3]{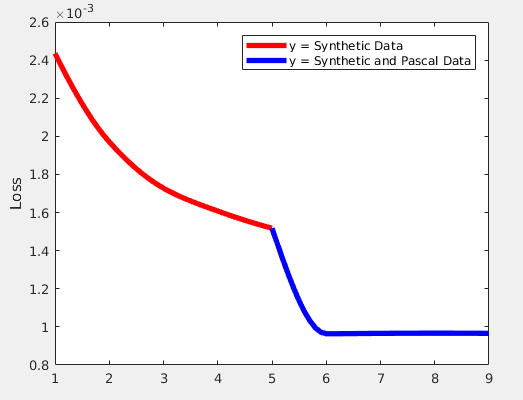}
	\caption{Progress of training the hourglass CNN using the proposed render pipeline. Loss per epoch when training on synthetic data followed by finetuning on a mix of real and synthetic images is illustrated here.}
	\label{fig:loss}
	\vspace{-0.5cm}
\end{figure}

\subsection{Render Pipeline}

\begin{table*}[!h]
	\centering
	\caption{Quantitative analysis of several variants of the proposed object-SLAM approach. Here, the best and worst object localization errors refer to the errors in the most accurately localized and the most erroneously localized object respectively. We also show the average object localization error and accumulated endpoint drifts, wherever applicable.}
	\label{table:main}
	\begin{adjustbox}{max width=\linewidth}
		\begin{tabular}{|c|c|c|c|c|c|c|c|c|}
			\hline
			\multirow{2}{*}{\textbf{Sequence ID}} &  \multirow{2}{*}{\textbf{Approach}} & \multirow{2}{*}{\textbf{Mode}} & \multirow{2}{*}{\textbf{\# Objects}} &  \multicolumn{3}{c|}{\textbf{Object localization Error (metres)}} & \multicolumn{2}{c|}{\textbf{Drift of trajectory }} \\
			\hhline{~~~~-----}
			& & & & \multicolumn{1}{c|}{\textbf{Best}} & \multicolumn{1}{c|}{\textbf{Worst}} & \multicolumn{1}{c|}{\textbf{Avg}} & \textbf{Z direction (in meters)} & \textbf{X direction (in meters)}
			\\
			\hline
			\multirow{4}{*}{1} & ORB-SLAM \cite{ORB}& - &\multirow{4}{*}{11}& 0.3162 & 4.9535 & 2.8539 & 4.3 & 0.9051\\
			\hhline{~--~-----}
			&Ours (without object loop closure)& batch & & 0.2526 & 2.7017 & 1.0318 & 2.051 & \textbf{0.18} \\
			\hhline{~--~-----}
			&\multirow{2}{*}{Ours (with object loop closure)}& batch & & 0.2193 & \textbf{1.6584} & \textbf{0.779} & 0.804 & 0.326 \\
			\hhline{~~-~-----}
			& & incremental & & \textbf{0.2188} & 1.6756 & 0.7998 &\textbf{ 0.79} & 0.209 \\
			\hline
			\multirow{4}{*}{2} & ORB-SLAM \cite{ORB} & - &\multirow{4}{*}{7}& 0.0607 & 0.8734 & 0.3734 & 0.22 & 0.6181\\
			\hhline{~--~-----}
			&Ours (without object loop closure)& batch & & 0.0791 & 0.7306 & 0.3125 & 0.19 & 0.4309 \\
			\hhline{~--~-----}
			&\multirow{2}{*}{Ours (with object loop closure)}& batch & & 0.0738 & 0.6005 & 0.2419 & \textbf{0.04} & \textbf{0.13} \\
			\hhline{~~-~-----}
			& & incremental & & \textbf{0.072} & \textbf{0.5936} & \textbf{0.2364} & \textbf{0.04} & \textbf{0.13} \\
			\hline
			\multirow{3}{*}{3} & ORB-SLAM \cite{ORB} & - &\multirow{3}{*}{9}&\multicolumn{5}{c|} {\textbf{ORB-SLAM Fails to initialize}}\\
			\hhline{~--~-----}
			&\multirow{2}{*}{Ours (with object loop closure)}& batch & & 0.0915 & \textbf{0.9998} & \textbf{0.521} & 0.0194 & \textbf{0.0556} \\
			\hhline{~~-~-----}
			&& incremental & & \textbf{0.0825} & 1.0174 & 0.5219& \textbf{0.0045} & 0.06 \\
			\hline
			\multirow{3}{*}{4} & ORB-SLAM \cite{ORB} & - &\multirow{3}{*}{7}& 0.1113 & 0.5993 & 0.3657 & 0.101 & N/A \\
			\hhline{~--~-----}
			&\multirow{2}{*}{Ours (without object loop closure)}& batch & & 0.0661 & 0.4688 & 0.3086 & \textbf{0.055} & N/A \\
			\hhline{~~-~-----}
			&& incremental & & \textbf{0.0614} & \textbf{0.4678} & \textbf{0.3051} & 0.059 & N/A \\
			\hline
		\end{tabular}
		
	\end{adjustbox}
	\label{tab:mainTable}
\end{table*}

\begin{figure*}[!h]
	\centering
	\minipage{0.20\textwidth}
	\subfloat[Seq 1 with OLC]{\includegraphics[width=\linewidth, height=7cm]{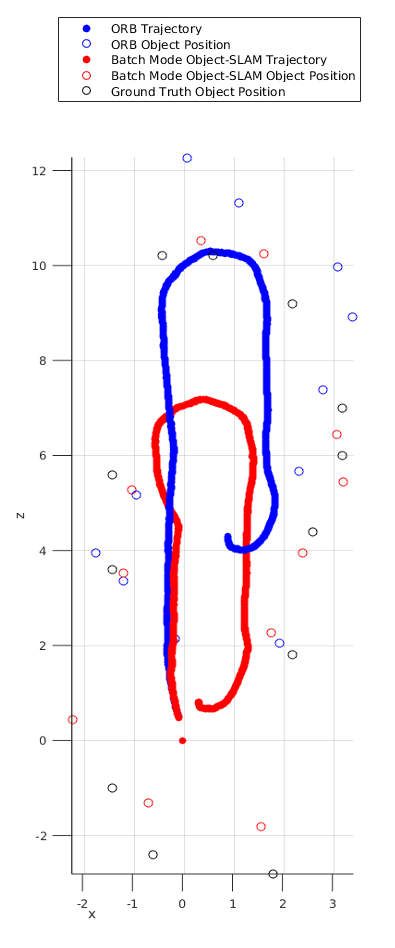}\label{fig:f1}}
	\endminipage  \hfill
	\minipage{0.20\textwidth}
	\subfloat[Seq 1 without OLC]{\includegraphics[width=\linewidth, height=7cm]{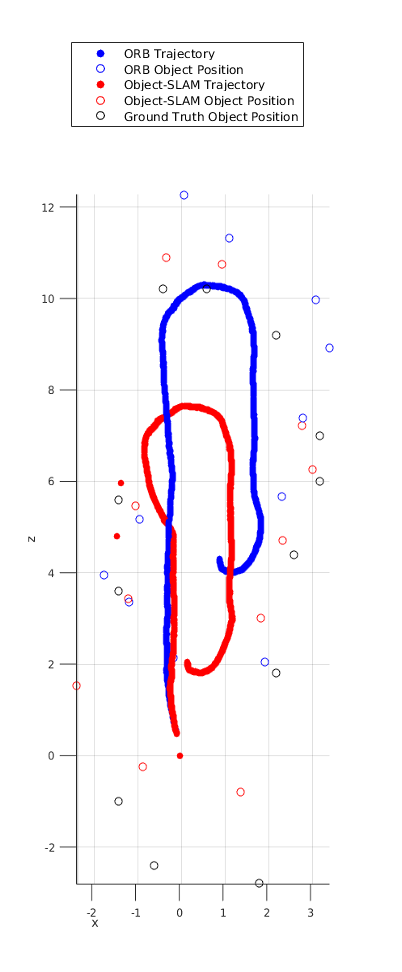}\label{fig:f2}}
	\endminipage \hfill
	\minipage{0.20\textwidth}
	\subfloat[Seq 2 with OLC]{\includegraphics[width=\linewidth, height=7cm]{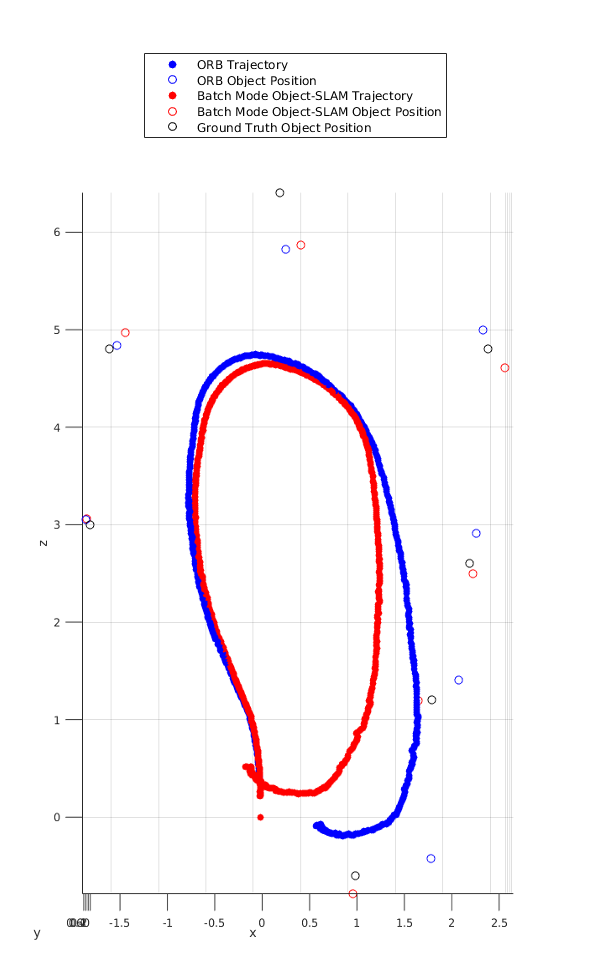}\label{fig:f3}}
	\endminipage  \hfill
	\minipage{0.20\textwidth}
	\subfloat[Seq 2 without OLC]{\includegraphics[width=\linewidth, height=7cm]{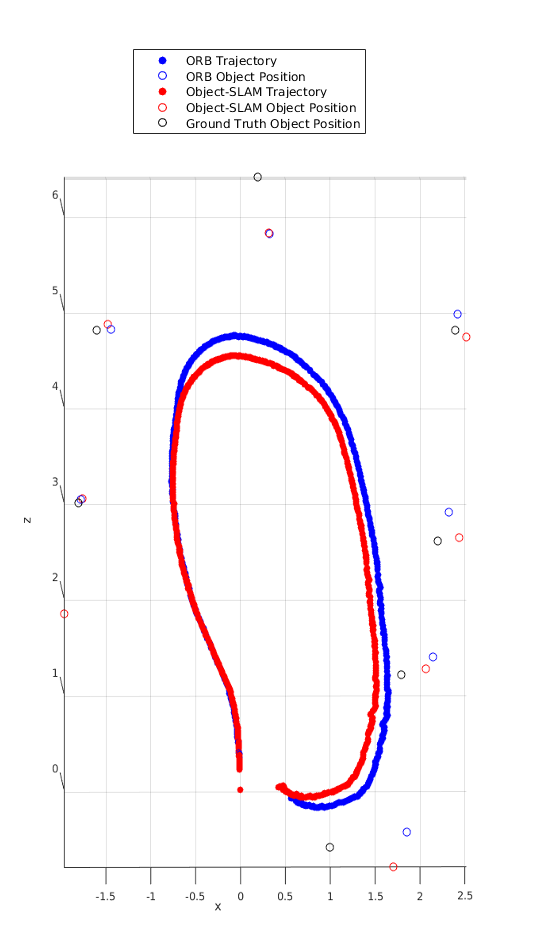}\label{fig:f1}}
	\endminipage  \hfill
	\minipage{0.20\textwidth}
	\subfloat[Seq 3 without OLC (top) and Seq 4 with OLC (bottom)]{\includegraphics[width=\linewidth, height=7cm]{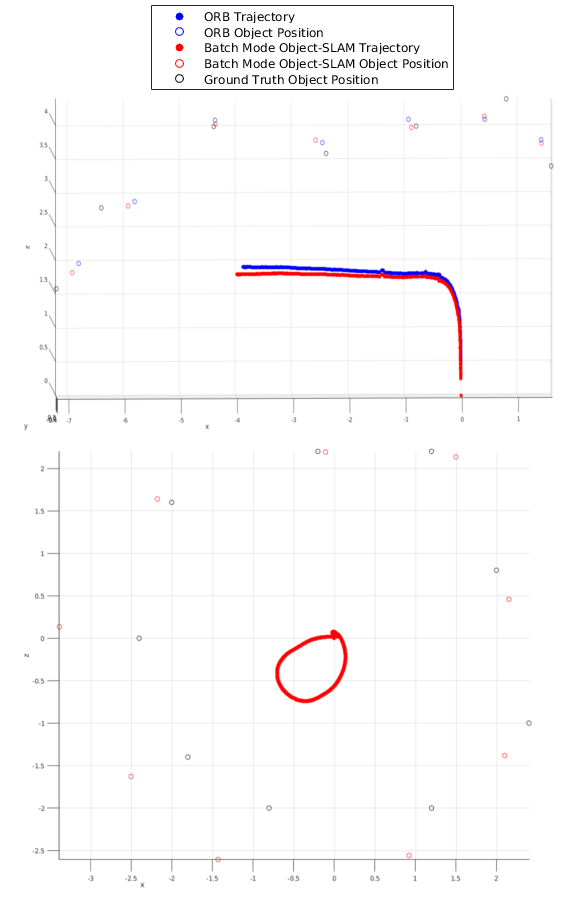}\label{fig:f2}}
	\endminipage\hfill
	\caption{ Estimated Trajectories and Object Locations by ORB-SLAM and Object-SLAM with/without \textbf{Object Loop Closure detection (OLC)} }
	\label{fig:trajectories}
\end{figure*}

For the task of training CNNs to detect keypoints for the chair category, we use our render pipeline to synthesize a little over a million keypoint annotated images of various models of chairs. We train the stacked hourglass \cite{hourglass,KM_IROS} architecture on this synthesized dataset. We then finetune the CNN on a smaller dataset that comprises of real chair images from the PASCAL3D+ \cite{PASCAL3D} dataset.

Keypoint localization accuracies for several configurations of the hourglass network are shown in TABLE \ref{table:keypointAccuracy}. We see that CNNs trained on synthetic data alone fail to generalize to real data from our sequences. Similarly, training on real data (from the PASCAL3D+ dataset) alone performs fairly well on test samples from the same dataset, but fails to generalize to other kinds of data, such as our sequence. However, finetuning the CNN on a combination of both real and synthetic data leads to better generalization. Fig. \ref{fig:loss} shows the decrease of loss with time when the best performing approach from TABLE \ref{table:keypointAccuracy} is trained using our render pipeline.
\vspace{-0.4cm}

\subsection{Discussion}
\vspace{-0.1cm}

On the collected dataset, we evaluate the several variants of the proposed object-SLAM approach, viz. batch mode (the factor graph is constructed for the entire scene and is optimized in one go), incremental mode (observations are processed as they arrive). We also evaluate object-SLAM with and without object-based loop closures. TABLE \ref{table:main} summarizes the results of these experiments.

For each of the approaches, we evaluate object localization error (in meters), and also the accumulated drift in the trajectory in the X and Z directions. The accumulated drift is computed only in cases where the robot returns to the starting position at the end of the run. Significant drifts in the Z-direction occur in sequence 1 for ORB-SLAM \cite{ORB}, but these are corrected by the object-SLAM system, regardless of the presence or absence of object loop closures. In sequence 4, there is a strong rotation (a perpendicular turn), where ORB-SLAM performs poorly.

Our discriminative features enjoy an advantage over tracked features \cite{ORB} in that they need to be visible in just a single image to enable object measurements. We demonstrate this in sequence 3 (cf. Fig. \ref{fig:circle}). In this sequence the robot purely rotates about its origin. While ORB-SLAM \cite{ORB} fails to initialize for this sequence, the object-SLAM proposed here is able to accurately localize all chairs in the scene. The circle traced out by our object-SLAM systems has a radius roughly equal to that of the drone used in the run.

The incremental version performs competitively, and in cases, better, than the batch-mode object-SLAM. Loop closures based on object observations further boost the accuracy of the proposed approach. For batch mode, we perform a chordal initialization before optimizing the factor graph.

Fig. \ref{fig:trajectories} shows plots of the trajectories estimated by ORB-SLAM \cite{ORB}, as well as from object-SLAM. We also show object location estimates overlaid on ground-truth object locations for the sequence. Sequences 1 and 2 are plotted with and without incorporating object loop closure constraints, to show that object loop closure further reduces endpoint drift.

\subsection{Instance Retrieval}

The proposed category-level model certainly achieves instance independence, but instance retrieval itself may be of use in several scenarios, such as Augmented Reality (AR). The shape parameters estimated by our approach can be used to guide an instance search over the keypoint annotated CAD collection. Given a query image (2D), we run a K-Nearest Neighbors search to retrieve the closest instance possible from the CAD collection. In Fig. \ref{fig:instanceRetrieval}, we present results from running a 5-nearest neighbor search and manually choosing the closest instance.

\begin{figure}[!tbh]
\centering
\includegraphics[width=0.75\linewidth]{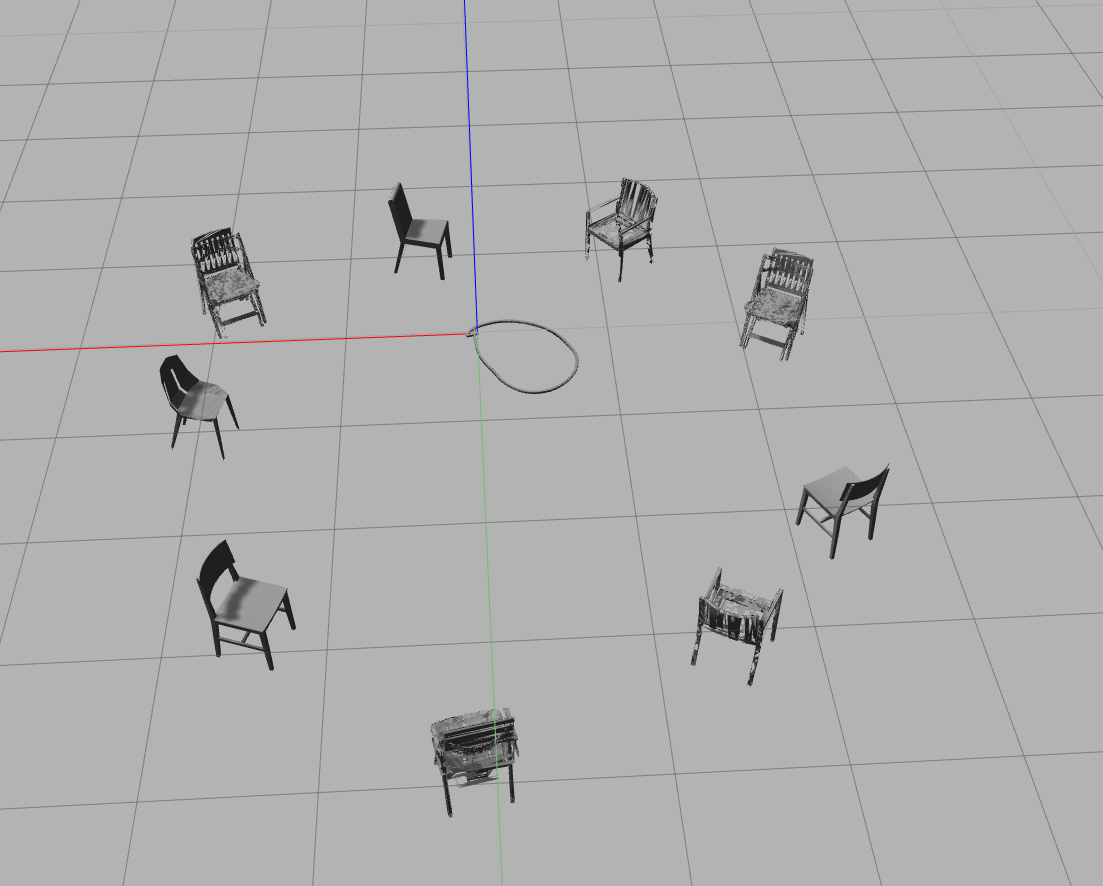}
\caption{Rendering objects estimated in a run where the robot rotates in place. ORB-SLAM \cite{ORB} fails to initialize due to insufficient parallax.}
\label{fig:circle}
\end{figure}

These instances can then be used to render the objects in the scene as well as the robot trajectory. One such rendering for Sequence 3 is shown in Fig \ref{fig:circle}.

\section{Conclusions}

\begin{figure}[!t]
	\centering
	\includegraphics[scale=0.5]{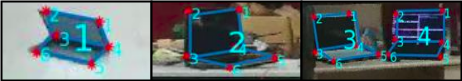}
	\caption{Qualitative results on another object category (laptops).}
	\vspace{-0.25cm}
\end{figure}

In this paper, we have presented an approach for monocular object-SLAM that relies on category-specific models, rather than relying on instance-specific models or generic models. This is a new paradigm in the nascent field of object-SLAM research. The proposed category-specific models can be adopted to many other such rigid objects, as long as data for the corresponding category is available in the form of aligned CAD models (See Fig. for an example). Future work could focus on further reducing the extent of supervision required to learn such category-specific models.

\section{Acknowledgements}

The authors acknowledge the support and funding from Kohli Center for Intelligent Systems (KCIS) IIIT Hyderabad, and Tata Consultancy Services (TCS) Innovation Labs India for this work. We also acknowledge the help of Akanksha Baranwal, Roopal Nahar, Danish Sodhi, Karnik Ram, and Sarthak Sharma for their timely assistance.


\bibliography{references}

\begin{thebibliography}{10}
\providecommand{\url}[1]{#1}
\csname url@rmstyle\endcsname
\providecommand{\newblock}{\relax}
\providecommand{\bibinfo}[2]{#2}
\providecommand\BIBentrySTDinterwordspacing{\spaceskip=0pt\relax}
\providecommand\BIBentryALTinterwordstretchfactor{4}
\providecommand\BIBentryALTinterwordspacing{\spaceskip=\fontdimen2\font plus
\BIBentryALTinterwordstretchfactor\fontdimen3\font minus
  \fontdimen4\font\relax}
\providecommand\BIBforeignlanguage[2]{{%
\expandafter\ifx\csname l@#1\endcsname\relax
\typeout{** WARNING: IEEEtran.bst: No hyphenation pattern has been}%
\typeout{** loaded for the language `#1'. Using the pattern for}%
\typeout{** the default language instead.}%
\else
\language=\csname l@#1\endcsname
\fi
#2}}

\bibitem{ORB}
M.~J. M.~M. Mur-Artal, Ra\'ul and J.~D. Tard\'os, ``{ORB-SLAM}: a versatile and
  accurate monocular {SLAM} system,'' \emph{IEEE Transactions on Robotics},
  vol.~31, no.~5, pp. 1147--1163, 2015.

\bibitem{LSD}
J.~Engel, T.~Sch{\"o}ps, and D.~Cremers, ``Lsd-slam: Large-scale direct
  monocular slam,'' in \emph{European Conference on Computer Vision}.\hskip 1em
  plus 0.5em minus 0.4em\relax Springer, 2014, pp. 834--849.

\bibitem{Sunderhauf}
\BIBentryALTinterwordspacing
N.~S{\"{u}}nderhauf, T.~Pham, Y.~Latif, M.~Milford, and I.~D. Reid,
  ``Meaningful maps - object-oriented semantic mapping,'' \emph{CoRR}, vol.
  abs/1609.07849, 2016. [Online]. Available:
  \url{http://arxiv.org/abs/1609.07849}
\BIBentrySTDinterwordspacing

\bibitem{SLAM++}
R.~F. Salas-Moreno, R.~A. Newcombe, H.~Strasdat, P.~H. Kelly, and A.~J.
  Davison, ``Slam++: Simultaneous localisation and mapping at the level of
  objects,'' in \emph{Proceedings of the IEEE conference on computer vision and
  pattern recognition}, 2013, pp. 1352--1359.

\bibitem{Paull}
B.~Mu, S.-Y. Liu, L.~Paull, J.~Leonard, and J.~P. How, ``Slam with objects
  using a nonparametric pose graph,'' in \emph{Intelligent Robots and Systems
  (IROS), 2016 IEEE/RSJ International Conference on}.\hskip 1em plus 0.5em
  minus 0.4em\relax IEEE, 2016, pp. 4602--4609.

\bibitem{Choudhary}
S.~Choudhary, L.~Carlone, C.~Nieto, J.~Rogers, Z.~Liu, H.~I. Christensen, and
  F.~Dellaert, ``Multi robot object-based slam,'' in \emph{International
  Symposium on Experimental Robotics}.\hskip 1em plus 0.5em minus 0.4em\relax
  Springer, 2016, pp. 729--741.

\bibitem{Sunderhauf2015}
N.~S{\"u}nderhauf, F.~Dayoub, S.~McMahon, M.~Eich, B.~Upcroft, and M.~Milford,
  ``Slam--quo vadis? in support of object oriented and semantic slam,'' 2015.

\bibitem{SfM++}
M.~Crocco, C.~Rubino, and A.~Del~Bue, ``Structure from motion with objects,''
  in \emph{Proceedings of the IEEE Conference on Computer Vision and Pattern
  Recognition}, 2016, pp. 4141--4149.

\bibitem{RAS2016}
D.~G{\'a}lvez-L{\'o}pez, M.~Salas, J.~D. Tard{\'o}s, and J.~Montiel,
  ``Real-time monocular object slam,'' \emph{Robotics and Autonomous Systems},
  vol.~75, pp. 435--449, 2016.

\bibitem{KM_ICRA}
J.~K. Murthy, G.~S. Krishna, F.~Chhaya, and K.~M. Krishna, ``Reconstructing
  vehicles from a single image: Shape priors for road scene understanding,'' in
  \emph{Proceedings of the IEEE Conference on Robotics and Automation}, 2017.

\bibitem{Zia}
M.~Z. Zia, M.~Stark, and K.~Schindler, ``Towards scene understanding with
  detailed 3d object representations,'' \emph{International Journal of Computer
  Vision}, vol. 112, no.~2, pp. 188--203, 2015.

\bibitem{tulsiani_PAMI}
S.~Tulsiani, A.~Kar, J.~Carreira, and J.~Malik, ``Learning category-specific
  deformable 3d models for object reconstruction.'' \emph{IEEE transactions on
  pattern analysis and machine intelligence}, 2016.

\bibitem{ISAM2}
M.~Kaess, H.~Johannsson, R.~Roberts, V.~Ila, J.~J. Leonard, and F.~Dellaert,
  ``isam2: Incremental smoothing and mapping using the bayes tree,'' \emph{The
  International Journal of Robotics Research}, vol.~31, no.~2, pp. 216--235,
  2012.

\bibitem{RiSE}
D.~M. Rosen, M.~Kaess, and J.~J. Leonard, ``Rise: An incremental trust-region
  method for robust online sparse least-squares estimation,'' \emph{IEEE
  Transactions on Robotics}, vol.~30, no.~5, pp. 1091--1108, 2014.

\bibitem{KM_IROS}
J.~K. Murthy, S.~Sharma, and M.~Krishna, ``Shape priors for real-time monocular
  object localization in dynamic environments,'' in \emph{Proceedings of the
  IEEE Conference on Intelligent Robots and Systems}, 2017.

\bibitem{kostas_ICRA}
G.~Pavlakos, X.~Zhou, A.~Chan, K.~Derpanis, and K.~Daniilidis, ``6-dof object
  pose from semantic keypoints,'' in \emph{Proceedings of the IEEE Conference
  on Robotics and Automation (In Press)}, 2017.

\bibitem{hourglass}
A.~Newell, K.~Yang, and J.~Deng, ``Stacked hourglass networks for human pose
  estimation,'' in \emph{European Conference on Computer Vision}.\hskip 1em
  plus 0.5em minus 0.4em\relax Springer, 2016, pp. 483--499.

\bibitem{RenderForCNN}
H.~Su, C.~R. Qi, Y.~Li, and L.~J. Guibas, ``Render for cnn: Viewpoint
  estimation in images using cnns trained with rendered 3d model views,'' in
  \emph{The IEEE International Conference on Computer Vision (ICCV)}, December
  2015.

\bibitem{SVO}
C.~Forster, Z.~Zhang, M.~Gassner, M.~Werlberger, and D.~Scaramuzza, ``Svo:
  Semidirect visual odometry for monocular and multicamera systems,''
  \emph{IEEE Transactions on Robotics}, vol.~33, no.~2, pp. 249--265, 2017.

\bibitem{g2o}
R.~K{\"u}mmerle, G.~Grisetti, H.~Strasdat, K.~Konolige, and W.~Burgard, ``g 2
  o: A general framework for graph optimization,'' in \emph{Robotics and
  Automation (ICRA), 2011 IEEE International Conference on}.\hskip 1em plus
  0.5em minus 0.4em\relax IEEE, 2011, pp. 3607--3613.

\bibitem{GTSAM}
F.~Dellaert \emph{et~al.}, ``Gtsam,'' \emph{URL: https://borg. cc. gatech.
  edu}, 2012.

\bibitem{YOLO}
J.~Redmon and A.~Farhadi, ``Yolo9000: Better, faster, stronger,'' \emph{arXiv
  preprint arXiv:1612.08242}, 2016.

\bibitem{Fast-RCNN}
R.~Girshick, ``Fast r-cnn,'' in \emph{International Conference on Computer
  Vision ({ICCV})}, 2015.

\bibitem{VpsKps}
S.~Tulsiani and J.~Malik, ``Viewpoints and keypoints,'' in \emph{2015 IEEE
  Conference on Computer Vision and Pattern Recognition (CVPR)}.\hskip 1em plus
  0.5em minus 0.4em\relax IEEE, 2015, pp. 1510--1519.

\bibitem{Shapenet}
A.~X. Chang, T.~Funkhouser, L.~Guibas, P.~Hanrahan, Q.~Huang, Z.~Li,
  S.~Savarese, M.~Savva, S.~Song, H.~Su, \emph{et~al.}, ``Shapenet: An
  information-rich 3d model repository,'' \emph{arXiv preprint
  arXiv:1512.03012}, 2015.

\bibitem{blender}
\BIBentryALTinterwordspacing
{Blender Online Community}, \emph{Blender - a 3D modelling and rendering
  package}, Blender Foundation, Blender Institute, Amsterdam, 2017. [Online].
  Available: \url{http://www.blender.org}
\BIBentrySTDinterwordspacing

\bibitem{PASCAL3D}
Y.~Xiang, R.~Mottaghi, and S.~Savarese, ``Beyond pascal: A benchmark for 3d
  object detection in the wild,'' in \emph{IEEE Winter Conference on
  Applications of Computer Vision (WACV)}, 2014.

\bibitem{chandraker2015}
S.~Song and M.~Chandraker, ``Joint sfm and detection cues for monocular 3d
  localization in road scenes,'' in \emph{Proceedings of the IEEE Conference on
  Computer Vision and Pattern Recognition}, 2015, pp. 3734--3742.

\bibitem{Junaid_ICRA}
J.~K. M. K. M.~K. Sarthak~Sharma, Junaid Ahmed~Ansari, ``Beyond pixels:
  Leveraging geometry and shape cues for online multi-object tracking,'' in
  \emph{Proceedings of the IEEE Conference on Robotics and Automation (In
  Press)}, 2018.

\bibitem{Ceres}
S.~Agarwal, K.~Mierle, and Others, ``Ceres solver,''
  \url{http://ceres-solver.org}.

\end{thebibliography}
\bibliographystyle{IEEEtran}	

\end{document}